\documentclass[12pt,a4paper,twoside]{article}
\usepackage[]{graphicx}
\usepackage{pstricks,pst-text,pst-node,pst-plot}
\usepackage{epsf}
\usepackage{amssymb}
\usepackage{amsmath}

\newenvironment{proof}[1][Proof]{\textbf{#1.} }{\
\rule{0.5em}{0.5em}}

\newcommand{\adj}{{\rm adj}}
\newcommand{\card}{{\rm card}}
\newcommand{\supp}{{\rm supp}}

\title{{\Huge Properties of the Discrete Pulse Transform for
Multi-Dimensional Arrays}}
\author{Roumen Anguelov and Inger Fabris-Rotelli\\
Department of Mathematics and Applied Mathematics\\
University of
Pretoria\\roumen.anguelov@up.ac.za\\inger.fabris-rotelli@up.ac.za
\vspace{3cm} \\ ISBN: 978-1-86854-785-2 \\ \\ Technical Report
\hspace{7cm} 2010/01}

\date{}
\begin{document}

\maketitle \newpage

\section{Introduction}

This report presents properties of the Discrete Pulse Transform on
multi-dimensional arrays introduced earlier in \cite{AngPlas}. The
main result given here in Lemma 2.1 is also formulated in
\cite[Lemma 21]{AngFab}. However, the proof, being too technical,
was omitted there and hence it appears in full in this publication.

\section{The Lemma}

The lemma which follows deals with two technical aspects of the
Discrete Pulse Transform of a function $f \in
\mathcal{A}(\mathbb{Z}^d)$ (where $\mathcal{A}(\mathbb{Z}^d)$
denotes a vector lattice). The first is that the Discrete Pulse
representation of a function $f$, given by
\begin{equation*} f = \sum_{n=1}^N D_n(f),
\end{equation*} can be written as the sum of individual pulses of
each resolution layer $D_n(f)$. The second result in the lemma below
indicates a form of linearity for the nonlinear LULU operators.

{\bf Lemma 2.1} \\ Let $f\in\mathcal{A}(\mathbb{Z}^d)$,
$\supp(f)<\infty$, be such that $f$ does not have local minimum sets
or local maximum sets of size smaller than $n$, for some $n \in
\mathbb{N}$. Then we have the following two results.
\begin{itemize}
\item[a)]\begin{eqnarray}\label{lemmaEq11}
(id-P_n)f=\sum_{i=1}^{\gamma^-(n)}\phi_{ni}+\sum_{j=1}^{\gamma^+(n)}\varphi_{nj},
\end{eqnarray} where $V_{ni}=\supp(\phi_{ni}), i=1,2,...,\gamma^-(n)$, are local minimum sets of $f$ of size
$n$, $W_{nj} = \supp(\varphi_{nj}),j=1,2,...,\gamma^+(n)$, are local
maximum sets of $f$ of size $n$,  $\phi_{ni}$ and $\varphi_{nj}$ are
negative and positive discrete pulses respectively, and we also have
that
\begin{eqnarray}
\bullet&&\hspace{-6mm} V_{ni} \cap V_{nj} = \emptyset  \textrm{ and
} \adj(V_{ni}) \cap V_{nj} = \emptyset,\ i,j =
1,...,\gamma^-(n),\ i\neq j,\hspace{1cm} \label{lemmaEq111}\\
\bullet&&\hspace{-6mm}W_{ni} \cap W_{nj} = \emptyset  \textrm{ and }
\adj(W_{ni}) \cap W_{nj} = \emptyset, \ i,j = 1,...,\gamma^+(n),
i\neq j, \label{lemmaEq112}\\
\bullet&&\hspace{-6mm}V_{ni} \cap W_{nj} = \emptyset \
i=1,...,\gamma^-(n)\ ,j = 1,...,\gamma^+(n).\label{lemmaEq113}
\end{eqnarray}
\item[b)] For every fully trend preserving operator $A$
\begin{eqnarray*}
U_n(id-AU_n)=U_n-AU_n,\\
L_n(id-AL_n)=L_n-AL_n.
\end{eqnarray*}
\end{itemize}

\begin{proof} \\
a) Let $V_{n1}, V_{n2},..., V_{n\gamma^-(n)}$ be all local minimum
sets of size $n$ of the function $f$. Since $f$ does not have local
minimum sets of size smaller than $n$, then $f$ is a constant on
each of these sets, by \cite[Theorem 14]{AngFab}. Hence, the sets
are disjoint, that is $V_{ni}\cap V_{nj}=\emptyset$, $i\neq j$.
Moreover, we also have
\begin{equation}\label{lemmaEq3}
\adj(V_{ni})\cap V_{nj}=\emptyset,\ i,j=1,...,\gamma^-(n).
\end{equation}
Indeed, let $x\in\adj(V_{ni})\cap V_{nj}$. Then there exists $y\in
V_{ni}$ such that $(x,y)\in r$. Hence $y\in V_{ni}\cap\adj(V_{nj})$.
From the local minimality of the sets $V_{ni}$ and $V_{nj}$ we
obtain respectively $f(y)<f(x)$ and $f(x)<f(y)$, which is clearly a
contradiction. For every $i=1,...,\gamma^-(n)$ denote by $y_{ni}$
the point in $\adj(V_{ni})$ such that
\begin{equation}\label{lemmaEq3a}
f(y_{ni})=\min\limits_{y\in\adj(V_{ni})}f(y).
\end{equation}
Then we have
\[
U_nf(x)=\left\{\begin{tabular}{cl}$f(y_{ni})$&if $x\in V_{ni}$,
$i=1,...,\gamma^-(n)$\\\\$f(x)$& otherwise (by \cite[Theorem
9]{AngFab})\end{tabular}\right.
\]
Therefore
\begin{equation}\label{thelemmaaeq1}
(id-U_n)f=\sum_{i=1}^{\gamma^-(n)}\phi_{ni}
\end{equation}
where $\phi_{ni}$ is a discrete pulse with support $V_{ni}$ and
negative value (down pulse).

Let $W_{n1},W_{n2},...,W_{n\gamma^+(n)}$ be all local maximum sets
of size $n$ of the function $U_nf$. By \cite[Theorem 12(b)]{AngFab}
every local maximum set of $U_nf$ contains a local maximum set of
$f$. Since $f$ does not have local maximum sets of size smaller than
$n$, this means that the sets $W_{nj}$, $j=1,...,\gamma^+(n)$, are
all local maximum sets of $f$ and $f$ is constant on each of them.
Similarly to the local minimum sets of $f$ considered above we have
$W_{ni}\cap W_{nj}=\emptyset$, $i\neq j$, and $\adj(W_{ni})\cap
W_{nj}=\emptyset$, $i,j=1,...,\gamma^+(n)$. Moreover, since $U_n(f)$
is constant on any of the sets $V_{ni}\cup\{y_{ni}\}$,
$i=1,...,\gamma^-(n)$, see \cite[Theorem 14]{AngFab}, we also have
\begin{equation}\label{lemmaEq4}
(V_{ni}\cup\{y_{ni}\})\cap W_{nj}=\emptyset,\ i=1,...,\gamma^-(n),\
j=1,...,\gamma^+(n),
\end{equation}
which implies (\ref{lemmaEq113}).

Further we have
\[
L_nU_nf(x)=\left\{\begin{tabular}{cl}$U_nf(z_{nj})$&if $x\in
W_{nj}$, $j=1,...,\gamma^+(n)$ \\\\ $U_nf(x)$&
otherwise\end{tabular}\right.
\]
where $z_{nj}\in\adj(W_{nj})$, $j=1,...,\gamma^+(n)$, are such that
$U_nf(z_{nj})=\max\limits_{z\in\adj(W_{nj})}U_nf(z)$. Hence
\begin{equation}\label{lemmaEq113i}
(id-L_n)U_nf=\sum_{j=1}^{\gamma^+(n)}\varphi_{nj}
\end{equation}
where $\varphi_{nj}$ is a discrete pulse with support $W_{nj}$ and
positive value (up pulse). \newline Thus we have shown that
\[
(id-P_n)f=(id-U_n)f+(id-L_n)U_nf=\sum_{i=1}^{\gamma^-(n)}\phi_{ni}+\sum_{j=1}^{\gamma^+(n)}\varphi_{nj}.
\]

b) Let the function $f\in\mathcal{A}(\mathbb{Z}^d)$ be such that it
does not have any local minimum or local maximum sets of size less
than $n$. Denote $g=(id-AU_n)(f)$. We have
\begin{equation}\label{theoLemmaeqb1}
g=(id-AU_n)(f)=(id-U_n)(f)+((id-A)U_n)(f).
\end{equation}
As in a) we have that (\ref{thelemmaaeq1}) holds, that is  we have
\begin{equation}\label{theoLemmaeqb3}
(id-U_n)(f)=\sum_{i=1}^{\gamma^-(n)}\phi_{ni},
\end{equation}
where the sets $V_{ni}=\supp(\phi_{ni})$, $i=1,...,\gamma^-(n)$, are
all the local minimum sets of $f$ of size $n$ and satisfy
(\ref{lemmaEq111}). Therefore
\begin{equation}\label{theoLemmaeqb2}
g=\sum_{i=1}^{\gamma^-(n)}\phi_{ni}+((id-A)U_n)(f).
\end{equation}
Furthermore,
\[
U_n(f)(x)=\left\{\begin{tabular}{lll}$f(x)$&if&$x\in\mathbb{Z}^d\setminus\bigcup\limits_{i=1}^{\gamma^-(n)}V_{ni}$
\\ \\ $v_i$&if&$x\in V_{ni}\cup\{y_{ni}\}$,
$i=1,...,\gamma^-(n)$,\end{tabular}\right.
\]
where $v_i=f(y_{ni})=\min\limits_{y\in\adj(V_{ni})}f(y)$. Using that
$A$ is fully trend preserving, for every $i=1,...,\gamma^-(n)$ there
exists $w_i$ such that $((id-A)U_n)(f)(x)=w_i$, $x\in
V_{ni}\cup\{y_{ni}\}$. Moreover, using that every adjacent point has
a neighbor in $V_{ni}$ we have that
$\min\limits_{y\in\adj(V_{ni})}((id-A)U_n)(f)(y)=w_i$. Considering
that the value of the pulse $\phi_{ni}$ is negative, we obtain
through the representation (\ref{theoLemmaeqb2}) that $V_{ni}$,
$i=1,...,\gamma^-(n)$, are local minimum sets of $g$.

Next we show that $g$ does not have any other local minimum sets of
size $n$ or less. Indeed, assume that $V_0$ is a local minimum set
of $g$ such that $\card(V_0)\leq n$. Since
$V_0\cup\adj(V_0)\subset\mathbb{Z}^d\setminus\bigcup\limits_{i=1}^{\gamma^-(n)}V_{ni}$
it follows from (\ref{theoLemmaeqb2}) that $V_0$ is a local minimum
set of $((id-A)U_n)(f)$. Then using that $(id-A)$ is neighbor trend
preserving and using \cite[Theorem 17]{AngFab} we obtain that there
exists a local minimum set $W_0$ of $U_n(f)$ such that $W_0\subseteq
V_0$. Then applying again \cite[Theorem 17]{AngFab} or \cite[Theorem
12]{AngFab} we obtain that there exists a local minimum set
$\tilde{W}_0$ of $f$ such that $\tilde{W}_0\subseteq W_0\subseteq
V_0$. This inclusion implies that $\card(\tilde{W}_0)\leq n$. Given
that $f$ does not have local minimum sets of size less than $n$ we
have $\card(\tilde{W}_0)= n$, that is $\tilde{W}_0$ is one of the
sets $V_{ni}$ - a contradiction. Therefore, $V_{ni}$,
$i=1,...,\gamma^-(n)$, are all the local minimum sets of $g$ of size
$n$ or less. Then using again (\ref{thelemmaaeq1}) we have
\begin{equation}\label{theoLemmaeqb4}
(id-U_n)(g)=\sum_{i=1}^{\gamma^-(n)}\phi_{ni}
\end{equation}
Using (\ref{theoLemmaeqb3}) and (\ref{theoLemmaeqb4}) we obtain
\[
(id-U_n)(g)=(id-U_n)(f)
\]
Therefore
\begin{eqnarray*}
(U_n(id-AU_n))(f)&=&U_n(g)\ = \ g-(id-U_n)(f)\\&=&
(id-AU_n)(f)-(id-U_n)(f)\\
&=& (U_n-AU_n)(f).
\end{eqnarray*}
This proves the first identity. The second one is proved in a
similar manner.
\end{proof}

\end{document}